\documentclass[runningheads]{llncs}
\usepackage{graphicx}

\usepackage{tikz}
\usepackage{comment}
\usepackage{amsmath,amssymb} 
\usepackage{color}
\usepackage{multirow}

\usepackage{xspace}
\newcommand*{\ie}{i.e.\@\xspace}
\newcommand*{\eg}{e.g.\@\xspace}
\newcommand*{\etal}{et al.\@\xspace}

\usepackage{kotex}
\newcommand{\first}[1]{\textcolor{red}{\textbf{#1}}}
\newcommand{\second}[1]{\textcolor{blue}{\textbf{#1}}}

\usepackage[accsupp]{axessibility}  

\begin{document}
\pagestyle{headings}
\mainmatter
\def\ECCVSubNumber{7893}  

\title{Non-Parametric Style Transfer} 

\titlerunning{Non-Parametric Style Transfer}
%
\author{Jeong-Sik Lee \and
Hyun-Chul Choi}
\authorrunning{J.-S. Lee, H.-C. Choi}
%
\institute{ICVSLab.,Department of Electronic Engineering, Yeungnam University, Gyeongsan, 38541 Republic of Korea\\
\email{\{a2819z, pogary\}@ynu.ac.kr}\\
\url{http://pogary.yu.ac.kr}}
\maketitle

\begin{abstract}
	Recent feed-forward neural methods of arbitrary image style transfer mainly utilized encoded feature map upto its second-order statistics, \ie, linearly transformed the encoded feature map of a content image to have the same mean and variance (or covariance) of a target style feature map. In this work, we extend the second-order statistical feature matching into a general distribution matching based on the understanding that style of an image is represented by the distribution of responses from receptive fields. For this generalization, first, we propose a new feature transform layer that exactly matches the feature map distribution of content image into that of target style image. Second, we analyze the recent style losses consistent with our new feature transform layer to train a decoder network which generates a style transferred image from the transformed feature map. Based on our experimental results, it is proven that the stylized images obtained with our method are more similar with the target style images in all existing style measures without losing content clearness. 
\end{abstract}

\section{Introduction}
Image style transfer has been improved in (1) generating speed of stylized image, (2) number of embedded styles in a network, and (3) quality of stylized image.
Sub-real time speed of style transfer regarding (1) has already been achieved by using feed-forward neural networks~\cite{Johnson_2016_ECCV,Ulyanov_2016_ICML}.
Multiple or arbitrary style transfer regarding (2) has also been achieved by utilizing a feature normalization layer as a linear style transform layer~\cite{Dumoulin_2017_ICLR,Huang_2017_ICCV,Li_2017_NIPS} which changes mean and variance (or covariance) of a feature map into those of target style.
In the aspect of output style quality regarding (3), some recent feed-forward neural methods utilized multiple style transform layers~\cite{Li_2017_NIPS,sheng2018avatar} and patch-based matching~\cite{Chen2016FastPS,sheng2018avatar} for an improved feature distribution matching.
In this trend of image style transfer, the previous methods was commonly utilizing linear transform layer(s) based on 2nd-order feature statistics. Even though some recent methods tried to match the distribution of a feature map from content image to target style image more accurately, they still did partial distribution matching.

Under the agreement to this trend of improvement, \ie, transferring whole impression for an image by matching the responses of receptive fields in statistical distribution, we propose a new arbitrary style transfer method which exactly matches the feature map distribution upto subtle style difference. In our method, first, we transforms the feature map distribution of content image into that of target style image by a new transform layer of correlation-aware histogram matching.
Second, we train decoder network to minimize reconstruction loss or style loss, \ie, distribution matching loss, which is a consistent style measure with our feature transform layer. Based on appropriate experiments, it will be proven that the stylized images obtained by using our method are closer to the target images in all existing style measures without losing content clearness. 
In brief, the contributions of this work are as follows:

1. Exact distribution matching layer: we utilizes a new transform layer, \ie, exact distribution matching (EDM) layer, instead of feature transform layer based on 2nd-order feature statistics. As the result, we allows our style transfer network to transfer a style exactly with regard to the distribution of impression.

2. Channel-correlation-awareness: encoded feature map has a large number of channels, and it is impossible to define the physical meaning between channels.
To deal with this channel-correlation, we present a method for matching multivariate distribution of feature map.

3. Distribution loss: For training decoder network, we analyze the previously presented losses, \ie, higher-order moment loss and distribution loss, that measures the difference in distribution between two feature maps of output style and target style. 
Among the presented losses, we experimentally show which loss is suitable for style transfer.

4. Higher style quality and style capacity: We experimentally proves that exact distribution matching scheme both qualitatively and quantitatively improves the output style of a style transfer network.

In the remained parts of this paper, we will review the previous works that are closely related to this work in sec.\ref{sec:related work}, describe what are and how to apply our distribution matching layer and loss in sec.\ref{sec:method}, present experimental results in sec.\ref{sec:experiments}, and conclude this work in sec.\ref{sec:conclusion}.

\section{Related work}
\label{sec:related work}

In various fields including style transfer, numerous studies have attempted to match the distributions of features with different domains.
In this section, we describe the previous style transfer methods in perspective of distribution matching.

\subsection{Moment Matching for Style Transfer}
Gatys \etal \cite{Gatys_2016_CVPR} proposed a style transfer method by minimizing the two gram matrices of content and style features.
Inspired that matching the gram matrix is equivalent to minimizing the MMD \cite{li2017demystifying}, Huang \etal \cite{Huang_2017_ICCV} proposed adaptive instance normalization (AdaIN), which matches the mean and standard deviation of each channel of a feature.
Li \etal \cite{Li_2017_NIPS} introduced a WCT transform layer that considers the correlation between channels of features.
WCT requires more resources than AdaIN because they used principal component analysis (PCA) to match the covariance.
However, the above methods only match up to the 2nd-order moment, so the pattern of the style image is not properly expressed in the result image.

To solve this problem, Kalischek \etal \cite{kalischek2021light} used the CMD as a style loss and performed style transfers by matching the higher-order moments.
However, this method only can be adopted for optimization-based style transfer such as Gatys \etal because there is no appropriate transform layer that matched the higher-order moment.

These moment matching-based style transfers can match up to the 2nd-order moment or even higher-order, but there is a limitation in that they cannot be used in real-time as an optimization-based method.

\subsection{Distribution Matching for Style Transfer}
Recently, in style transfer, matching the distribution itself has been conducted \cite{sheng2018avatar,mroueh2019wasserstein,risser2020optimal}.
To match the distribution more closely, Sheng \etal \cite{sheng2018avatar} employed both patch-based style decorator and 2nd-order moment matching (AdaIN) at multi-scale.
Mroueh \etal \cite{mroueh2019wasserstein} assumed the distribution of content and style features followed the Gaussian distribution and matched the two distributions using the closed-form of the optimal transport for Gaussian distribution.
However, they could not perfectly match the distribution because they still assumed that the feature follow the Gaussian distribution.
Risser \etal \cite{risser2020optimal} proposed a hybrid model that combines the advantages of optimization-based and feed-forward method for texture synthesis and style transfer.
To exactly match the distribution, they introduced the sliced histogram matching that project source and target distributions to random orthogonal bases and match the 1-dimensional marginal distribution.
However, because this sliced histogram matching has to be repeated enough, it is slower than a general feed-forward network.
Heitzh \etal \cite{heitz2021sliced} employed the sliced Wasserstein distaance (SWD) as a style loss instead of gram matrix and achieved better stylistic outcomes than Gatys \etal. However it is still an optimization-based approach that is hard to operate in real-time.
Although these methods could better match the two feature distributions, they still require optimization-based or hybrid models and this results in a slow-speed style transfer due to the absence of a proper transform layer.

\section{Method}
\label{sec:method}

\subsection{Non-parametric Style Transfer: Exact Distribution Matching (EDM) layer}

Given an encoded feature map, $F \in \mathbb{R}^{C \times (H \times W)}$, from encoder network of $C$ convolution channels, $H$ height, and $W$ width, we can assume that this feature map is a sample set which consists of $H \times W$ points of $C$-dimensional random variable and image style transfer can be considered as matching the distributions of two different sample sets for their underlying probability density function (PDF) of the random variable. For instance, AdaIN \cite{Huang_2017_ICCV} (or WCT \cite{Li_2017_NIPS}) matches mean and variance (or covariance) under the assumption of Gaussian PDF to exactly match the distribution of the feature map.

There are two main problems in realizing our exact distribution matching (EDM). First, we need to exactly match the distribution of an encoded feature map to that of a target feature map without any given underlying PDF. Second, we have to deal with multi-variate distribution of which random variable consists of correlated channels. In this section, we will focus on these two problems and describe our solution for them.

\subsubsection{Exact Histogram Specification}

In general case, when underlying PDF is not given, matching exact distribution of two different sample sets needs a complex procedure beyond linear transformation such as AdaIN \cite{Huang_2017_ICCV} and WCT \cite{Li_2017_NIPS}. Histogram specification (HS) \cite{Hall_1974,Zhang_1992} is a well known PDF matching technique for this purpose in image processing area. Given two sets of scalar valued pixels, \ie, source and target sets, histogram specification is done as sorting each set in ascending order of pixel values followed by overwriting the source sorted values with the target sorted values. However, once there are several pixels of the same value, the sorted order cannot differentiate these pixels and this results in an inexact PDF matching. Exact histogram specification (EHS) \cite{Coltuc_2006} solved this problem by using spatial filters of various size and sorting of the filter responses giving priority to the response of smaller filter. We adopt EHS for our non-parametric style transform layer.

Assuming two single-channeled feature maps $F_C,\:F_S\in \mathbb{R}^{1 \times (H \times W)}$ of content image $I_C$ and style image $I_S$, we firstly obtain a sorted sample set $f_S \in \mathbb{R}^{1 \times (H \cdot W)}$ of $F_S$ in ascending order of channel value. Secondly, we obtain an extended feature map $\hat F_C \in \mathbb{R}^{K \times (H \times W)}$ which has K-dimensional responses of $F_C$ corresponding to pre-defined $K$ spatial filters. Then, a sorted 2-dimensional indices $x_C \in \mathbb{N}^{2 \times (H \cdot W)}$ of $\hat F_C$ is obtained by lexicographical sorting in ascending order of K-dimensional responses giving priority to the dimension corresponding to smaller filter. Finally, we can obtain a style-transformed feature map $F_{C|S} \in \mathbb{R}^{1 \times (H \times W)}$ by rearranging the sorted sample set $f_S$ according to the sorted order $x_C$ as eq.\ref{eq:histogram_specification}.
\begin{equation}
\label{eq:histogram_specification}
F_{C|S}(x_C(i)) = f_S(i)\: for\: 0 \leq i < H \times W.
\end{equation}

For sorting with priorities in filter responses, we design $K$ spatial filters with the following conditions.
A filter response that is highly associated with the current pixel should be given a high priority, while a filter response with weak correlation should be given a low priority.
We employ 10 neighborhood filters that fit these conditions, \ie, identity mapping and 9 filters in size of $3 \times 3$, $5 \times 5$, and $7 \times 7$. (Please refer to the supplementary material for detail).

Equation \ref{eq:histogram_specification} is successful in exactly matching single-channeled feature map distribution. But we need to deal with multi-channeled feature map in image style transfer. This issue will be dealt in the following section.

\subsubsection{Dealing with Multi-variate Distribution}

In color transfer, histogram specification on color image in Lab color space \cite{Grundland2005ColorHS} enables the color transformation without color distortion.
In the case of style transfer, we need to do histogram specification of multi-variate distributions with very large dimensions. In the encoded feature map, however, there is a correlation between channel whose physical meaning is unclear in the encoded feature map, unlike the lab color space where the physical meaning of each channels is clear and independent. 
AdaIN \cite{Huang_2017_ICCV}, which transforms multi-variate distribution without considering such channel-correlation, has a problem that the color of the content remained or the pattern of the style image did not appear properly.
WCT \cite{Li_2017_NIPS}, on the other hand, performs channel normalization to remove channel-correlation, then matching the two normalized features and channel-correlating again, allowing for style transfer without the artifacts found in AdaIN.

We employ the covariance matching approach of WCT to conduct histogram matching while taking channel-correlation into account in our EDM layer.

\subsubsection{Exact Distribution Matching}
\begin{figure}[t]\centering
    \includegraphics[width=\linewidth]{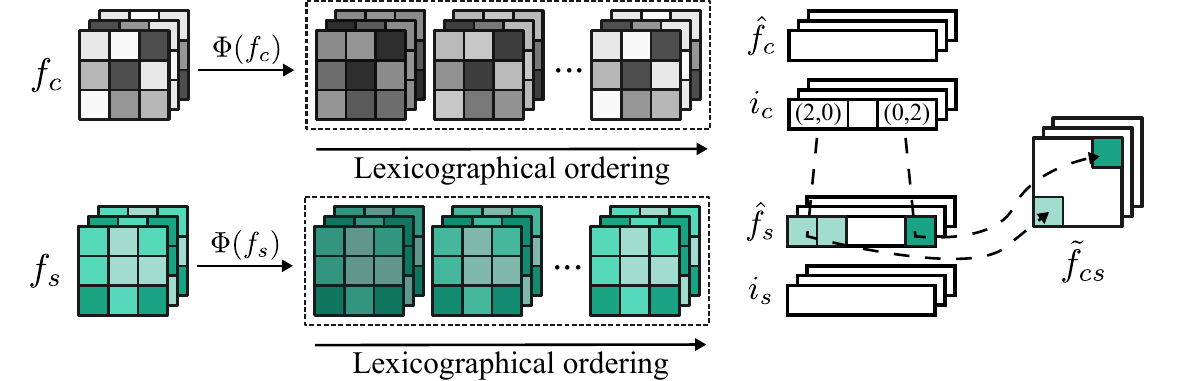}
    \caption{Overview of EHS. For simplicity, only a part of sorted values ($\hat{f}_c,\hat{f}_s$) and sorted indices ($i_c,i_s$) are shown, and the value of the sorted index denotes (row, col) of matrix.}
    \label{fig:ehs}
\end{figure}
EDM deals with the channel correlation by matching the covariance of the channel-wise histogram-matched feature ($\tilde{f}_{cs}$) and the style feature ($f_s$), and the whole matching process is as follows:

\textbf{Step 1. Channel-wise histogram matching}
Firstly, we channel-wisely match the histogram of the content feature ($f_c$) to the histogram of the style feature ($f_s$) using EHS.
To do matrix operation manner, we define $K$ neighborhood filters as one filter vector $\Phi$:
\begin{equation}\label{eq:filter_vector}
    \Phi = (\phi_1, \phi_2, \cdots, \phi_K)
\end{equation}
Thanks to this vector definition, we can get the two filter response ($\Phi(f_c), \Phi(f_s) \in \mathbb{R}^{K\times N\times HW}$) by doing convolution between filter vector ($\Phi$, eq.\ref{eq:filter_vector}) and content feature ($f_c$) or style feature ($f_s$):
\begin{equation}\label{eq:filter_response}
    \Phi(f_c) = \Phi \ast f_c, \quad \Phi(f_s) = \Phi \ast f_s,
\end{equation}
where $\ast$ is convolution operator.
By performing lexicographic ordering (eq.\ref{eq:lexicographic_ordering}) with $K$ filter responses (eq.\ref{eq:filter_response}) for each of the two features, the two ordered features ($\hat{f}_c, \hat{f}_s$) and ordering indices ($i_c, i_s$) are obtained.
\begin{equation}\label{eq:lexicographic_ordering}
    [\hat{f}_c, i_c] = \text{L-sort}(\Phi(f_c)), \quad [\hat{f}_s, i_s] = \text{L-sort}(\Phi(f_s)),
\end{equation}
where $\text{L-sort}(\cdot )$ means lexicographic ordering.
Finally, to match the distribution of two features, the value of the ordered style feature ($\hat{f}_s$) is rearranged according to the ordering index ($i_c$) of the content feature.
\begin{equation}\label{eq:ehs}
    \tilde{f}_{cs} = \text{EHS}(f_c, f_s) = \hat{f}_s [i_c],
\end{equation}
where $[\cdot]$ denotes tensor indexing.
By channel-wisely changing the value of the content feature to the value of the style features, we obtain a histogram-matched feature ($\tilde{f}_{cs}$), which denotes that the distribution itself is matched, and style transfer is completed.
Conversely, AdaIN assumed that each channel is a random variable following a Gaussian distribution and modified the features. However, our method is non-parametric distribution matching without a Gaussian distribution assumption, so the distribution can be matched more precisely and superior style transfer results can be obtained experimentally.

\textbf{Step 2. Channel-correlating}
To consider channel-correlation in the channel-wisely matched feature ($\tilde{f}_{cs}$), we perform covariance matching.
We normalize the mean and covariance of the histogram matched feature ($\tilde{f}_{cs}$) (eq.\ref{eq:whitening}), and match the normalized feature with the mean and covariance of the style feature once again (eq.\ref{eq:coloring}) to complete the feature transform considering channel-correlation.
\begin{equation}\label{eq:whitening}
    \bar{f}_{cs} = W_{cs} \otimes (\tilde{f}_{cs} - \mu(\tilde{f}_{cs})),
\end{equation}
\begin{equation}\label{eq:coloring}
    f_{cs} = C_{s} \otimes \bar{f}_{cs} + \mu (f_s),
\end{equation}
where $\otimes$ denotes matrix multiplication, and $W_{cs}$ and $C_{s}$ are whitening and coloring kernels respectively that derived from PCA like Li \etal \cite{Li_2017_NIPS}.

\begin{figure*}[t]\centering
    \includegraphics[width=\linewidth]{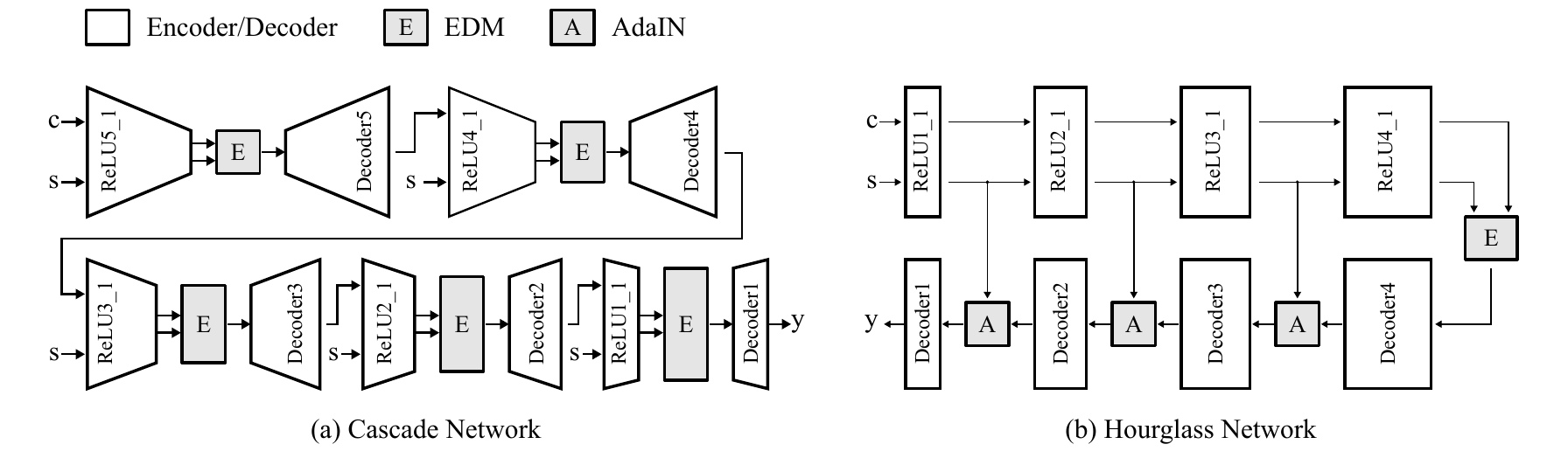}
    \caption{Overview of our multi-scale style transfer network. (a) For cascade network, the decoders are trained by reconstruction loss, then fix them. (b) For the hourglass network, the decoder trained by the content and style loss.}
    \label{fig:multi-scale-network}
\end{figure*}

\subsection{Multi-scaled Style Transfer}

Several methods \cite{Li_2017_NIPS,sheng2018avatar} introduced the multi-scaled style transfer architecture, \ie, cascade network, hourglass network, to more precisely match the distribution of the content feature and style feature.

To match the distribution at multi-scale, we adopt the cascade network \cite{Li_2017_NIPS} trained to reconstruct the input image, and apply the proposed EDM layer to each bottleneck.
However, in the EDM layer, EHS (eq.\ref{eq:ehs}) should sort tensors with the size of $HW$ as many as the number of filters ($K$), resulting in a complexity of $O(K\times HW\log(HW))$. Additionally, EDM performs PCA for channel-correlating, so the computational cost is relatively high.

To reduce the computational cost, we adopt the hourglass network \cite{sheng2018avatar} that only uses the EDM in the highest level and AdaIN \cite{Huang_2017_ICCV}(eq.\ref{eq:adain}), much lighter than EDM, in the other levels.
\begin{equation}\label{eq:adain}
    \text{AdaIN}(f_c, f_s) = \sigma (f_s) \left( \frac{f_c - \mu (f_c)}{\sigma (f_c)} \right) + \mu (f_s)
\end{equation}
where $\mu(\cdot)$ and $\sigma(\cdot)$ denote mean and standard deviation respectively.

\subsection{Non-parametric Style Loss: Uncorrelated distribution loss}\label{sec:style_loss}

Our hourglass network (EDM-Hourglass) is faster than our cascade network (EDM-Cascade), but poorer performance because of AdaIN that only matches up to 2nd-order moment.
To reduce the performance gap between the two models, we train the decoder of hourglass network using both content and style loss rather than using reconstruction loss.
In the remainder of current section, for training the hourglass network, we analyze the style losses presented in previous works and adopt one based on the analysis.

Existing style losses are limited by the transform layer, so only the lower-order moment is considered \cite{Huang_2017_ICCV}.
Some losses \cite{kalischek2021light,heitz2021sliced}, on the other side, can measure the higher-order moment or distribution itself, but they can only empolyed in optimization-based approaches because there is no corresponding transform layer.
Our method (EDM) that matches the distribution itself is not limited to the order of moment, so the previously proposed higher-order moment loss (CMD)\cite{kalischek2021light} and distribution loss (SWD)\cite{heitz2021sliced} can be adopted as a style loss for our hourglass network.

CMD allowed to match not only the 2nd-order moment but also the higher-order moment.
However, because it calculates moment channel-by-channel, the channel-correlation cannot be taken into account during the loss computation.
In addition, because computing the moment of infinite degree is impossible, the algorithm must be approximated and the distribution cannot be accurately matched due to the approximation.

The Wasserstein distance provides powerful geometric properties in comparing two distributions.
For 1-dimensional distribution, the Wasserstein distance has a closed-form solution (eq.\ref{eq:sw1d}), and SWD utilizes this property to define the distance between multi-variate distributions.
\begin{equation}\label{eq:sw1d}
    \mathcal{L}_{SW1D}(x, y) = \left \| \text{sort}(x) - \text{sort}(y)  \right \|^2.
\end{equation}
Firstly, SWD projects the n-dimensional source distribution and target distribution to the random direction ($V\in \mathbb{S}^M$) of the m-dimensional unit sphere ($\mathbb{S}^M$)
After projecting, we get the distance using eq.\ref{eq:sw1d} for 2 sets of $M$ 1-dimensional marginal distribution and define the average of these $M$ distances as the distance of two distributions (eq.\ref{eq:swd}).
\begin{equation}\label{eq:swd}
    \mathcal{L}_{SWD}(X, Y) = \frac{1}{M} \sum_{v\in V} \mathcal{L}_{SW1D}(X_v, Y_v),
\end{equation}
where $X_v$ is dot product between feature $X$ and random direction $v$.

SWD can converge faster as the dimension of the unit sphere is larger\cite{heitz2021sliced}. And the distance between two distributions can be measured implicitly considering the channel-correlation.
Therefore, we employ the SWD as a style loss that can compute the distance between distributions considering channel-correlation.
And in the aspect of computational cost, we set the dimension of the unit sphere to 4 times the dimension of the feature, \ie, the dimension of the unit sphere will be 2048 if the dimension of the feature is 512.

\section{Experiments}
\label{sec:experiments}

\subsection{Experimental Setup}
We get the content feature from \{relu3\_1\} (content layer, $L_c$) and the style features from \{relu1\_1, relu2\_1, relu3\_1, relu4\_1, relu5\_1\} (style layers, $L_s$) of VGG-19.
According to the aforementioned settings, the cascade network (fig.\ref{fig:multi-scale-network}(a)) consists of 4 encoders that use up to each layer of style layers ($L_s$) except for relu5\_1, and 4 decoders that mirror each encoder structure.
The encoder of the hourglass network (fig.\ref{fig:multi-scale-network}(b)) employs up to relu4\_1 of VGG-19, and the decoder has a skip-connection in each level for using AdaIN.

For the training dataset, We employed MS-COCO train2014 \cite{Lin_2014_ECCV} as content image and Painter By Number \cite{painter_by_numbers} as style image, and each dataset has about 80,000 images.
We resized the image to 512 size while maintaining the aspect ratio, and then randomly cropped to $256\times 256$.
We used the Adam optimizer with a learning rate of $10^{-4}$, and set the batch size to 8 and epoch 4 for training.

Each decoder of the cascade network was trained using a reconstruction loss (eq.\ref{eq:reconstruction}) so that the result image ($y$) that has passed through the encoder-decoder only, except the transform layer (EDM), recovers the input content image ($x_c$) as it is.
\begin{equation}\label{eq:reconstruction}
    \mathcal{L}_{recon} = \left|| y-x_c |\right|^2_2 + \frac{1}{|L_s|} \sum_{F\in L_s}\left|F(y) - F(x_c)  |\right|_2^2
\end{equation}
where $|\cdot |$ denotes the cardinality of a set.

For training the hourglass network, we followed the training methods proposed in the previous study \cite{Huang_2017_ICCV}, and the loss is equal to eq.\ref{eq:hourglass_loss}.
\begin{equation}\label{eq:hourglass_loss}
    \mathcal{L}_{total} = \mathcal{L}_{c} + \lambda_s \mathcal{L}_{s}
\end{equation}
where $\lambda_s$ is the weight of style loss, we set this value to 200.
We used the SWD (eq.\ref{eq:swd} as the style loss (eq.\ref{eq:style_loss}) and calculated by extracting the features of the stylized image ($y$) and style image ($x_s$) from the style layers ($L_s$).
\begin{equation}\label{eq:style_loss}
    \mathcal{L}_s = \frac{1}{|L_s|} \sum_{F\in L_s} \mathcal{L}_{SWD}( F(x_s), F(y))
\end{equation}

For numerical evaluation of our method, we used MS-COCO test2014 as the content and Painter By Number test as the style image respectively.

All of the experiments were conducted following environments: Pytorch framework v1.9.1, CUDA v11.3, CuDNN v8.2.0, and NVIDIA RTX 3090 GPU.

\subsection{Spatial filters for Exact Histogram Matching}

\begin{figure*}[t]\centering
    \includegraphics[width=\linewidth]{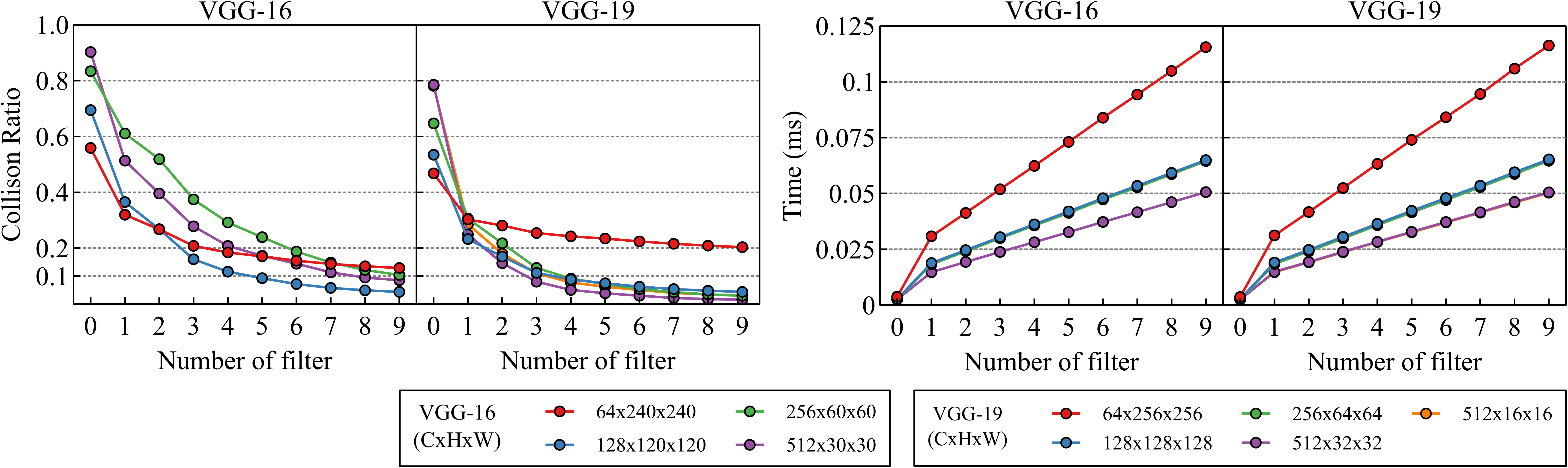}
    \caption{Bench for number of spatial filter. 0 on the x-axis denotes the result of using general sort, and 1\~9 denote the results of sorting using filters ranging from $3\times 3$ to a maximum of $7\times 7$.}
    \label{fig:exact_sort_bench}
\end{figure*}

\begin{figure}[t]\centering
    \includegraphics[width=\linewidth]{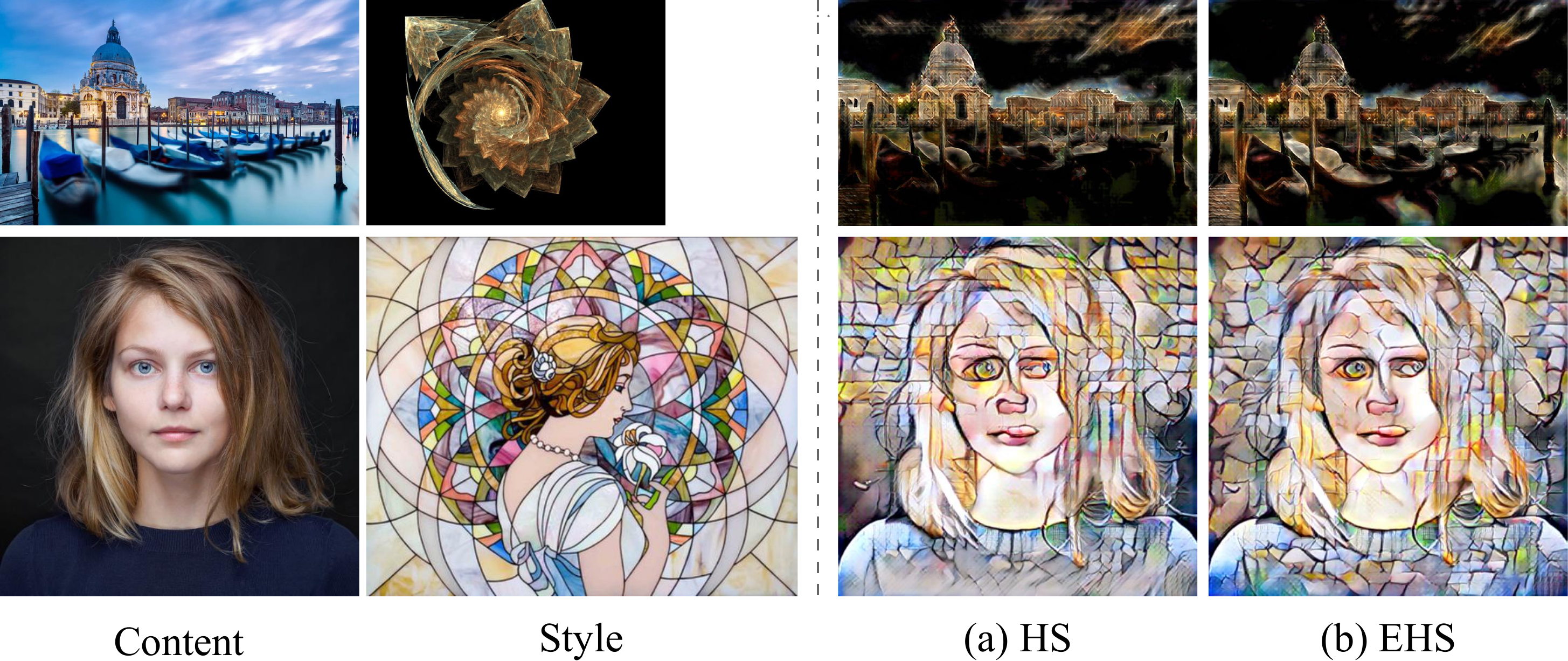}
    \caption{Ablation study about EHS. (a) Result of style transfer using HS, (b) Result of EHS using 9 filters. In the case of using EHS, the style pattern is uniformly distributed in the result.}
    \label{fig:exact_sort_ablation}
\end{figure}

Because our method uses lexicographical ordering, there is a trade-off in the speed of ordering and the collision ratio of values as the number of filters increases.
In order to determine the appropriate number of filters, we experiment with the speed and collision ratio according to the number of filters and show results in fig.\ref{fig:exact_sort_bench}
In fig.\ref{fig:exact_sort_bench}, the collision ratio represents the ratio of the number of pixels with the same value, and time represents the average of sorting time per feature.
(Please refer to the supplementary material for the data used in the experiment, the exact measurement method, the structure of each network, and how to calculate the collision ratio).

The left side of fig.\ref{fig:exact_sort_bench} shows the collision ratio according to the number of filters. We notice that the collision ratio reduces as the number of filters increases.
Since VGG-19 has a lower collision ratio than VGG-16 at most levels, we adopt the VGG-19 as an encoder for the style transfer network.

In terms of time consumption, as shown as the right side of fig.\ref{fig:exact_sort_bench}, since each filter response is sequentially used in the ordering process, the execution time increases according to the number of filters.
However, even with the largest feature resolution ($64\times 240\times 240$ and $64\times 256\times 256$), it still shows a reasonable execution speed in millisecond. Therefore, we select all filters that showed the lowest collision ratio to exactly specify the histogram.

Additionally, to identify the effect of EHS, we present the style transfer results with HS or EHS in fig.\ref{fig:exact_sort_ablation}.
In the case of HS (fig.\ref{fig:exact_sort_ablation}(a)), the patterns of style image are not evenly distributed but concentrated at the top of the image.
Especially, when HS is used in the first row (1st row of fig.\ref{fig:exact_sort_ablation}), the outlines of the ship in the lower part have disappeared, because the bright pattern of the style image is located in the upper part of the content image.
In contrast, those with EHS (fig.\ref{fig:exact_sort_ablation}(b)), style pattern is evenly distributed in the image.

\subsection{Comparison on Output Style Quality}

\begin{figure*}[t]\centering
    \includegraphics[width=\linewidth]{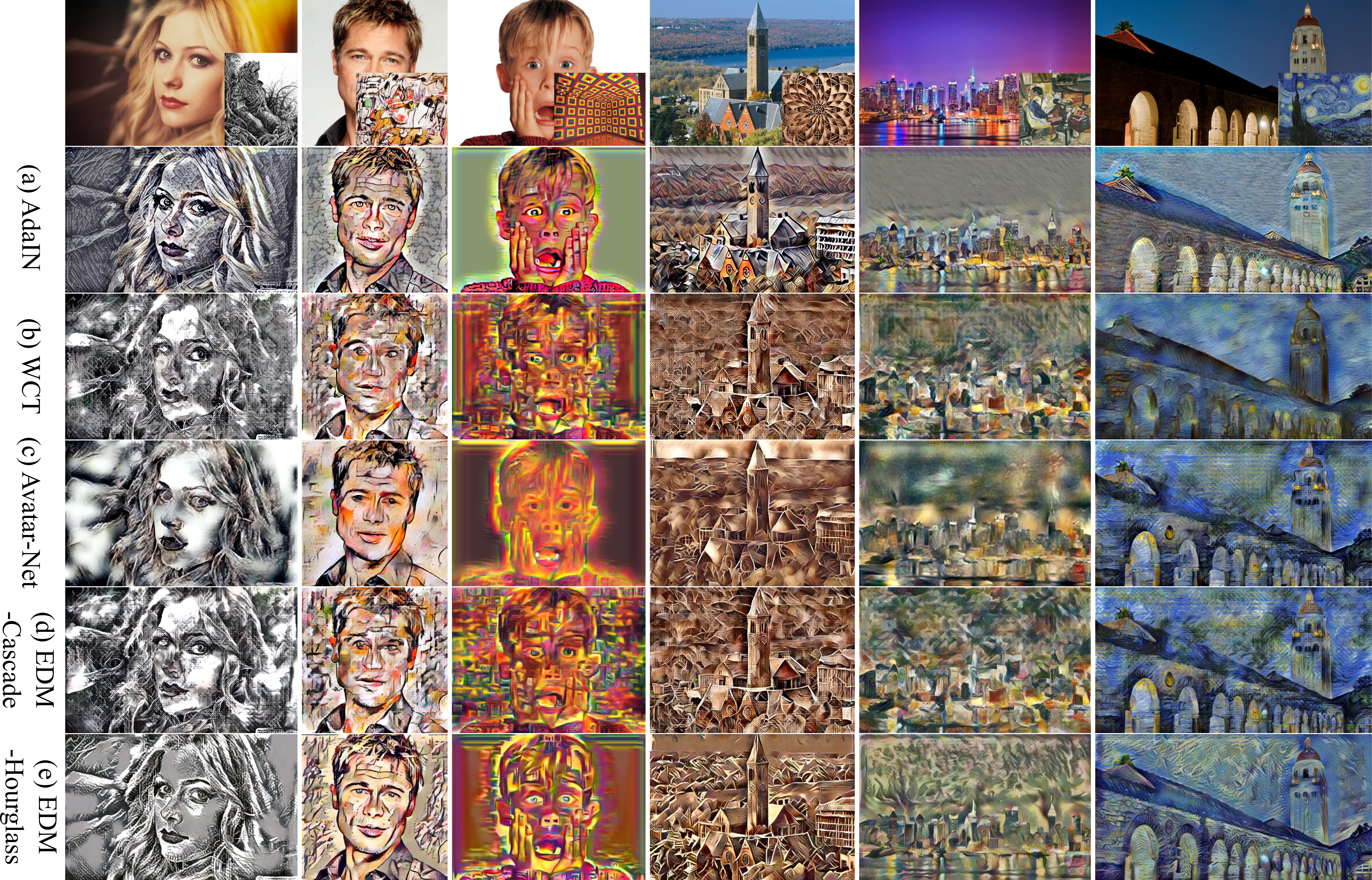}
    \caption{Comparison with other state-of-art methods. Style images shown in this figure are not used in training.}
    \label{fig:comparison}
\end{figure*}

\begin{table*}[t]\centering
    \begin{tabular}{|c|c|c|c|c|c|c|}\hline
         &  AdaIN \cite{Huang_2017_ICCV} & WCT \cite{Li_2017_NIPS} & Decorator \cite{sheng2018avatar} & HS & EHS & EDM \\ \hline
      content   & \first{2.2520} & 2.6591 & 3.6662 & 2.4017 & 2.2430 & \second{2.5930} \\
      gram \cite{Gatys_2016_CVPR} &  0.5980 & 0.5329 & \first{0.4532} & 0.5761 & 0.5768 & \second{0.4684} \\
      mean+std \cite{Huang_2017_ICCV} &  0.2923 & 0.2943 & \second{0.2701} & 0.2964 & 0.2970 & \first{0.2583} \\
      histogram-L2  &  0.4062 & 0.4085 & \second{0.3676} & 0.3992 & 0.4002 & \first{0.3586} \\
      CMD \cite{zellinger2017central} &  0.6393 & 0.6518 & \first{0.5690} & 0.6271 & 0.6204 & \second{0.6162} \\
      SWD \cite{heitz2021sliced} &  0.1845 & 0.1842 &  0.1896 & \second{0.1784} & 0.1803 & \first{0.1585} \\ \hline
    \end{tabular}
    \caption{Content and Style losses per transform layer. Red and blue text indicate the lowest and second lowest values respectively. histogram-L2 means the L2 distance of the weight values of the two histograms.}
    \label{tab:loss_table}
\end{table*}

\subsubsection{Qualitative Comparison.}
We present the style transfer results in fig.\ref{fig:comparison} to compare the proposed method and the state-of-the-art methods of arbitrary style transfer: AdaIN \cite{Huang_2017_ICCV}, WCT \cite{Li_2017_NIPS} and Avatar-Net \cite{sheng2018avatar}.
In the 1st, 2nd, and 4th columns in fig.\ref{fig:comparison}, EDM-Cascade shows a more clear boundary between background and foreground because of better contrast, and the pattern of style is uniformly placed in the result.
Most importantly, in the 3rd column in fig.\ref{fig:comparison}, WCT and Avatar-Net produce the artifacts in the background, conversely, EDM-Cascade shows the style pattern nicely without these artifacts. 
EDM-Hourglass also has artifacts in the background but clearly reflects the style pattern on the face compare to AdaIN and Avatar-Net.
Furthermore, other methods in the 5th and 6th columns in fig.\ref{fig:comparison} cannot express the brush texture of the style image in the sky part where there is no distinctive pattern in the content image, but the EDM series can.

\subsubsection{Quantitative Comparison.}
We also measured the content/style loss of each method to prove numerically that the proposed EDM performs better stylization than other transform layers.
In order to minimize the effect of loss used in learning and network structure, we calculate the loss by only changing the transform layer in the bottleneck of cascade network \cite{Li_2017_NIPS} and show in table \ref{tab:loss_table}.
In the case of the decorator, it has a low overall style loss, notably in CMD, but it is difficult to utilize in practice since memory consumption is quite large in large size features, \ie, actually it was unable to infer this model in a single GPU (VRAM 24GB), so we did it in the CPU.
Our method yields the lowest value in most of the style losses having a low content loss. CMD that has the lowest loss with decorator has the second-lowest value with our method.

\subsection{Analysis of style loss for training}
In this section, we investigate the effects of the loss, \ie, CMD \cite{kalischek2021light} and SWD \cite{heitz2021sliced}, on EDM-Hourglass training and analyze them in terms of visual quality and loss.
CMD and SWD have some hyper-parameters: one for order-term (p) how many orders to match, the other for the dimension of projection. 
The dimension of projection is determined by multiplying specific value (s) to the dimension of the feature.

\subsubsection{Perspective of Style Transfer Quality.}
\begin{figure*}[t]\centering
    \includegraphics[width=.8\linewidth]{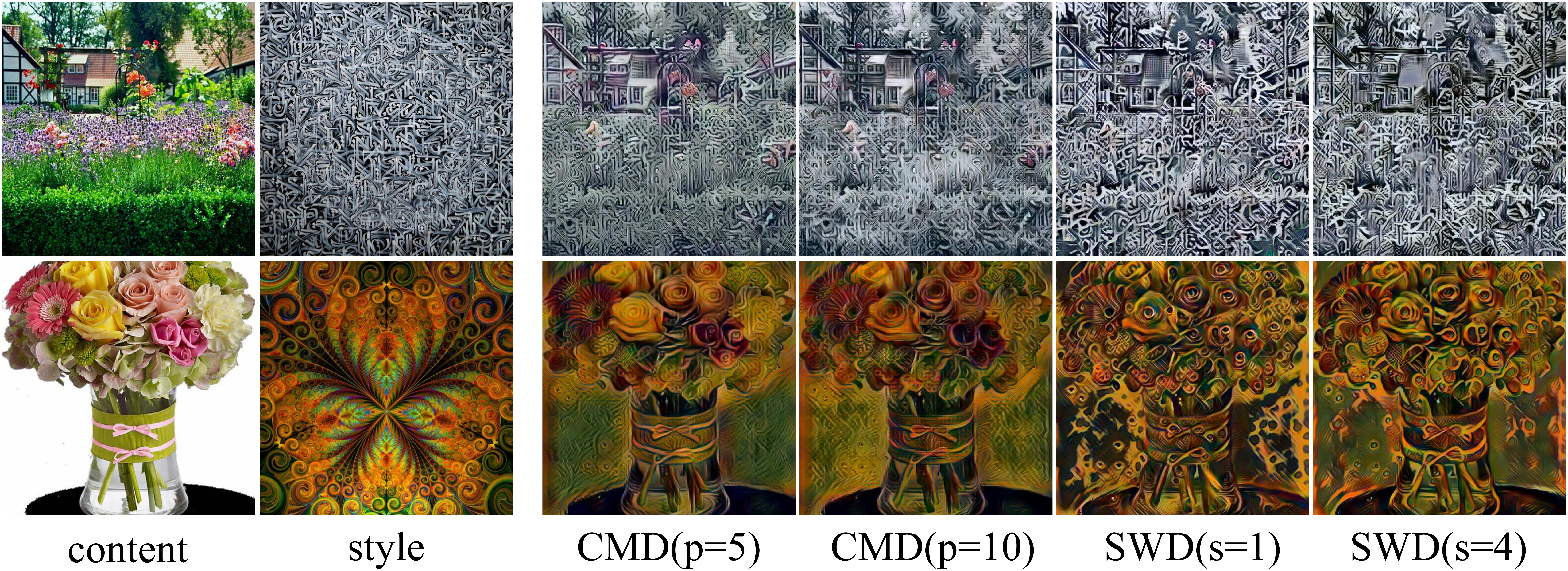}
    \caption{Visual comparison of CMD and SWD.}
    \label{fig:cmd_vs_swd}
\end{figure*}

Figure \ref{fig:cmd_vs_swd} shows the results of the network trained while changing the order of CMD and the projection dimension of SWD.
In the 1st row in fig.\ref{fig:cmd_vs_swd}, in the case of CMD, the pattern of the style image does not appear correctly and the green color of the content image remains.
In contrast, a network trained with SWD can generate an image that has proper pattern and color.
Specifically, in the 2nd row in fig.\ref{fig:cmd_vs_swd}, SWD clearly depicts the circle and curve patterns of the style better than CMD.
We reported the content and style losses in the supplemental material.

\subsection{Result of User Control}
To validate the flexibility of the proposed method, we show user controlling as in the previous studies \cite{Huang_2017_ICCV}, \ie, style strength control and multi-style interpolation.
We test user control with the cascade network because the cascade network is superior to the hourglass network and because our EDM can be used in all cascaded networks.

\begin{figure*}[t]\centering
    \includegraphics[width=\linewidth]{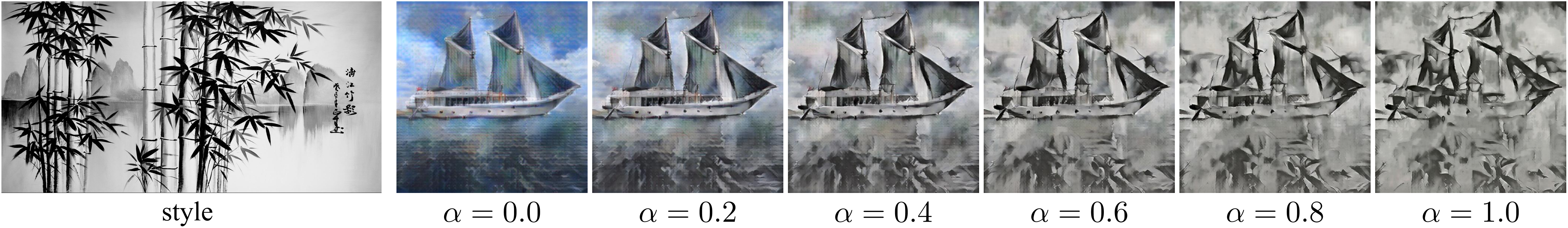}
    \caption{Style strength control.}
    \label{fig:style_strength}
\end{figure*}

\begin{figure*}[t]\centering
    \includegraphics[width=\linewidth]{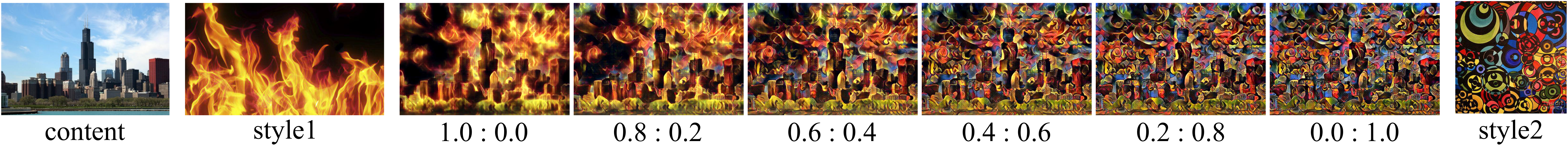}
    \caption{Interpolation between two styles. The ratio at the bottom of the result represents the weight of each style.}
    \label{fig:interpolation}
\end{figure*}

\subsubsection{Control style strength.}
The EDM can control the style strength by linear interpolating (eq.\ref{eq:style_strength})\cite{Huang_2017_ICCV} between the stylized feature ($f_{cs}$) and the content feature ($f_c$).
\begin{equation}\label{eq:style_strength}
    f_{cs} = \alpha f_{cs} + (1-\alpha )x_c
\end{equation}
Figure \ref{fig:style_strength} shows the output image by changing the $\alpha$ to [0, 0.2, 0.4, 0.6, 0.8, 1]. As style strength changes, the trade-off between content and style appears naturally.

\subsubsection{Multi-style interpolation.}
To interpolate between style images, we used the weighted sum (eq.\ref{eq:interpolation})\cite{Huang_2017_ICCV} between stylized features ($\text{EDM}(f_c, f_s^i)$). We show the results of style interpolation in fig.\ref{fig:interpolation}.
\begin{equation}\label{eq:interpolation}
    f_{cs} = \sum w_i \text{EDM}(f_c, f_s^i),
\end{equation}
where $\sum w_i=1$.
As shown in fig.\ref{fig:interpolation}, as the weight increases from style1 to style2, the fire pattern gradually changes to a circle pattern, and finally, the pattern of style2 is perfectly displayed in the image.

\subsection{Comparison on Computational Complexity}
\begin{table}[h]\centering
    \begin{tabular}{c|c|c}\hline
        \multirow{2}{*}{Method} & \multicolumn{2}{c}{Execution time (sec / image)} \\ \cline{2-3}
                                          & $256\times 256$ & $512\times 512$  \\ \hline
        AdaIN \cite{Huang_2017_ICCV}      & 0.003           & 0.004 \\
        WCT \cite{Li_2017_NIPS}           & 0.149           & 0.181 \\
        Avatar-Net \cite{sheng2018avatar} & 0.095           & 0.178\\ \hline
        EDM-Cascade                       & 0.158           & 0.220 \\
        EDM-Hourglass                     & 0.051           & 0.069 \\ \hline
    \end{tabular}
    \caption{Comparison execution time.}
    \label{tab:exe_time_comparision}
\end{table}
Table \ref{tab:exe_time_comparision} shows the average execution time for 1000 content/style image pairs.
Because EHS and PCA are used at all levels, EDM-Cascade is a bit slower than WCT, \ie, EDM-Cascade for each resolution is slower than 0.009 sec and 0.019 sec WCT, but there is no significant difference.
Compared to Avatar-Net, EDM-Hourglass shows a quicker speed, especially at $512\times 512$ resolution, \ie, EDM-Hourglass is 0.109 seconds faster than Avatar-Net, which is a significant speed difference.
This is because the decorator of Avatar-Net needs the higher the computational cost at high resolution. This result shows that our method is efficient for high-resolution images compared Avatar-Net

\section{Discussion and Conclusion}
\label{sec:conclusion}
In this work, we proposed a novel transform layer that matches the two distributions by considering the channel-correlation.
Existing studies mostly assumed that the encoded features follow a Gaussian distribution, but our EDM performs non-parametric distribution matching that matches the histogram itself without this assumption.

The proposed method is similar to the decorator of Avatar-Net in that it change the value of the source feature to the value of the corresponding target feature (histogram matching).
Since the decorator exchanges two patches with a high similarity, an odd pattern was appear in the result if there is no properly matching patch.
In contrast, EDM is free from this problem because it exchanges values based on the entire image rather than on a patch.
We experimentally showed that our method of changing values based on the entire image can distribute style patterns evenly and apply style patterns without artifacts in images with strong contrast, \eg, portrait image, between the background and objects compared to other methods including Avatar-Net.

The proposed EDM requires several filters to exactly specify the histogram, so we designed an appropriate filter and suggested the appropriate number of filters by analyzing the execution speed and collision ratio according to the number of filters.
In addition, we analyzed each transform layer by changing only the transform layer in the cascade network and showed that EDM performed the best style transfer with the lowest value in most style losses.
In the experiment of decorator in cascade network, we noted that decorator consumes very large memory as the size of feature increases, and found it is difficult to apply all layers of cascade network or hourglass network.

Additionally, among the recently introduced style losses, we experimentally demonstrated that SWD is appropriate as the style loss of the feed-forward network.
This style loss was utilized for style loss to reduce the performance gap between the hourglass and cascade networks. However, utilizing a transform layer that perfectly matched the distribution (EDM) to the decoder trained with the reconstruction loss had still yielded superior results.

As a future work, it is necessary to study a transform layer that can immediately consider channel-correlation when matching distribution or a feature encoding method without channel correlation.




%
%
\bibliographystyle{splncs04}
\bibliography{egbib}
\end{document}


\pagestyle{headings}
\mainmatter
\def\ECCVSubNumber{7893}  

\title{Supplementary Material for\\ Non-Parametric Style Transfer} 

\titlerunning{Non-Parametric Style Transfer}
%
\author{Jeong-Sik Lee \and
Hyun-Chul Choi
}
\authorrunning{J.-S. Lee, H.-C. Choi}
%
\institute{ICVSLab.,Department of Electronic Engineering, Yeungnam University, Gyeongsan, 38541 Republic of Korea\\
\email{\{a2819z, pogary\}@ynu.ac.kr}\\
\url{http://pogary.yu.ac.kr}}
\maketitle

\section{Spatial Filter Setup}

The 10 neighborhood filters we used for exact histogram specification (EHS) are listed below.

\begin{align*}
    &\phi_1 = \begin{bmatrix}1\end{bmatrix}~
    \phi_2 = \frac{1}{5}\begin{bmatrix}0&1&0\\1&1&1\\0&1&0\end{bmatrix}~
    \phi_3 = \frac{1}{9}\begin{bmatrix}1&1&1\\1&1&1\\1&1&1\end{bmatrix}~
    \phi_4 = \frac{1}{13}\begin{bmatrix}0&0&1&0&0\\0&1&1&1&0\\1&1&1&1&1\\0&1&1&1&0\\0&0&1&0&0\end{bmatrix}\\
    &\phi_5 = \frac{1}{21}\begin{bmatrix}0&1&1&1&0\\1&1&1&1&1\\1&1&1&1&1\\1&1&1&1&1\\0&1&1&1&0\end{bmatrix}~
    \phi_6 = \frac{1}{25}\begin{bmatrix}1&1&1&1&1\\1&1&1&1&1\\1&1&1&1&1\\1&1&1&1&1\\1&1&1&1&1\end{bmatrix}~
    \phi_7 = \frac{1}{25}\begin{bmatrix}0&0&0&1&0&0&0\\0&0&1&1&1&0&0\\0&1&1&1&1&1&0\\1&1&1&1&1&1&1\\0&1&1&1&1&1&0\\0&0&1&1&1&0&0\\0&0&0&1&0&0&0\end{bmatrix}\\
    &\phi_8 = \frac{1}{37}\begin{bmatrix}0&0&1&1&1&0&0\\0&1&1&1&1&1&0\\1&1&1&1&1&1&1\\1&1&1&1&1&1&1\\1&1&1&1&1&1&1\\0&1&1&1&1&1&0\\0&0&1&1&1&0&0\end{bmatrix}~
    \phi_9 = \frac{1}{45}\begin{bmatrix}0&1&1&1&1&1&0\\1&1&1&1&1&1&1\\1&1&1&1&1&1&1\\1&1&1&1&1&1&1\\1&1&1&1&1&1&1\\1&1&1&1&1&1&1\\0&1&1&1&1&1&0\end{bmatrix}~
    \phi_{10} = \frac{1}{49}\begin{bmatrix}1&1&1&1&1&1&1\\1&1&1&1&1&1&1\\1&1&1&1&1&1&1\\1&1&1&1&1&1&1\\1&1&1&1&1&1&1\\1&1&1&1&1&1&1\\1&1&1&1&1&1&1\end{bmatrix}~
\end{align*}

\section{Experiment setup of EHS analysis}

In the spatial filter experiment in Section 4.2 of the main text, we measured the complexity (time) and performance (collision ratio) of EHS using two networks (VGG-16 and VGG-19).
In this section, we describe in detail the network setup and measurement method in that experiment.

\subsubsection{Network Architecture.}
In the case of VGG-16, we extract the feature from \{relu1\_2, relu2\_2, relu3\_3, relu4\_2\} and in the case of VGG-19, from \{relu1\_1, relu2\_1, relu3\_1, relu4\_1, relu5\_1\}.
\subsubsection{Dataset}
For the test data, we used 1000 images of MS-COCO test2014 and cropped the images to the size of $240\times 240$ and $256\times 256$ for VGG-16 and VGG-19 respectively.
\subsubsection{How to calculate the collision ratio.}
Since our EHS channel-wisely matches the histogram, we calculated the collision ratio for each channel.
When only identity mapping ($\phi_1$) was performed, the collision ratio was simply calculated as the ratio of duplicate values per channel of the feature.
When the number of filters is 2 or more ($\Phi=\{\phi_1, \cdots \phi_K|K\ge 2\}$), we do lexicographical sorting using from 1 to $K$ filter responses ($F\in \mathbb{R}^{C\times HW}$) to get $K$ sorted value tensors and sorted index tensors.
Then we performed the linear searching on the $HW$ values of the $K$ index tensors. For two given tensors, if all the $K$ values of one tensor are the same as those of the other tensor, we judged that these two tensors are duplicated.

\section{Analysis of style loss for training perspective of style loss}

In addition to the visual quality comparison according to the style loss in the main text, we compared the style loss values.
All style losses tend to decrease as the ability to accurately measure the distance between the distributions of training losses (CMD and SWD) improves, \eg, $p:5\rightarrow 10$ and $s:1 \rightarrow 4$.
The network trained with SWD shows low style loss for all losses while the network trained with CMD shows low style loss only in CMD.
Based on this result, SWD is better than CMD as a loss for style transfer.

\begin{table}[h]\centering
    \begin{tabular}{|c|c|c|c|c|}\hline
                     & CMD (p=5) & CMD(p=10)& SWD (s=1) & SWD (s=4) \\ \hline
        content      & \first{5.1646}    & \second{5.1794}   & 6.8525    & 6.9016 \\
        gram         & 0.8234    & 0.8188   & \second{0.7474}    & \first{0.7392}\\
        mean+std     & 0.5127    & \second{0.5043}   & 0.5081    & \first{0.4874} \\
        histogram-L2 & 0.6391    & \second{0.6296}   & 0.6424    & \first{0.6191}\\
        CMD (p=10)   & \second{0.7765}    & \first{0.7739}& 0.7923    & 0.7849 \\
        SWD (s=4)    & 0.3151    & 0.3114   & \second{0.3103}    & \first{0.2972} \\ \hline
    \end{tabular}
    \caption{Ablation of style loss.}
    \label{tab:style_loss_ablation}
\end{table}

\clearpage
\section{Additional qualitative comparison}
\begin{figure}[h]\centering
    \includegraphics[width=\linewidth]{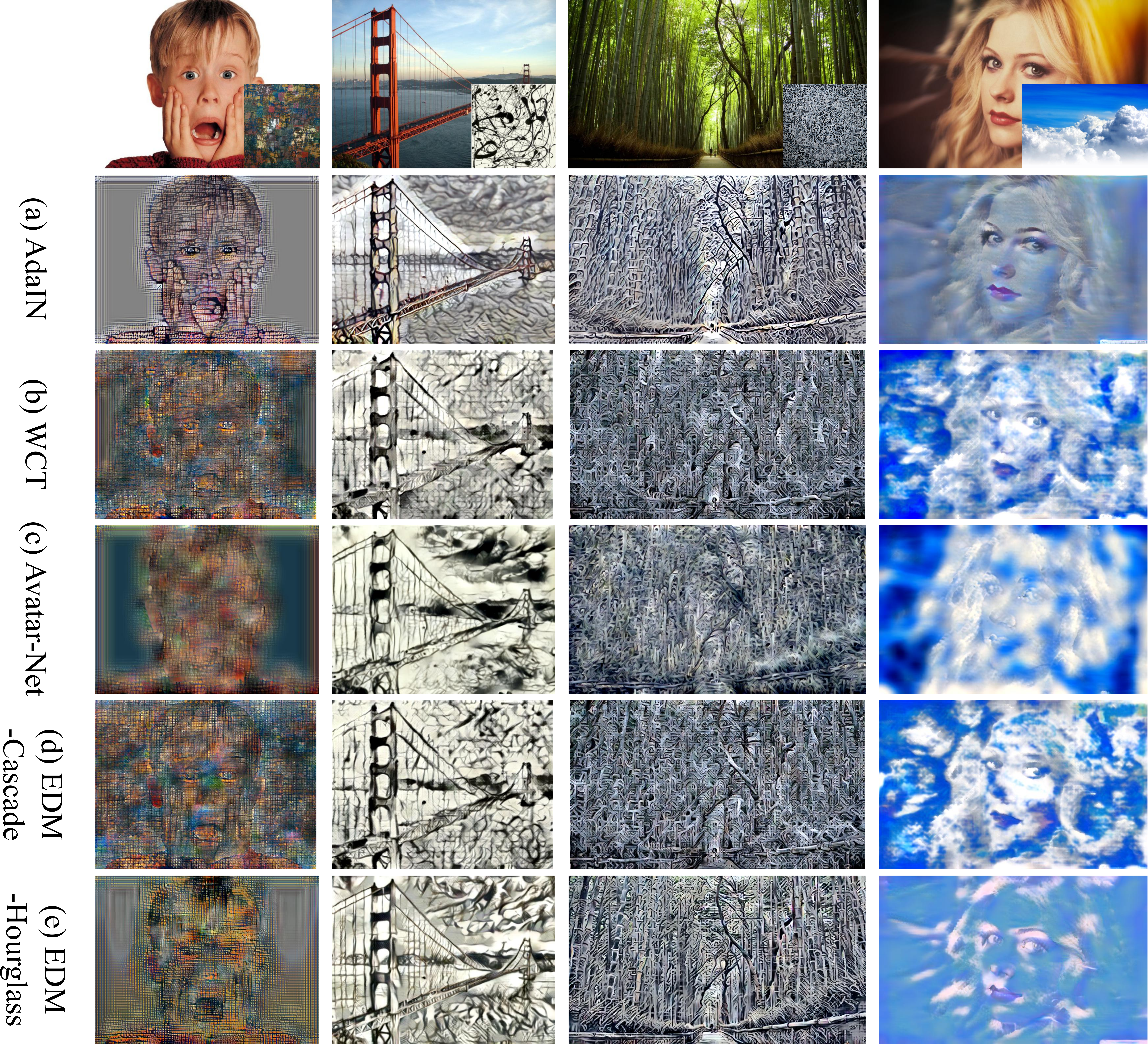}
    \caption{Additional comparison with other state-of-the-art: AdaIN\cite{Huang_2017_ICCV}, WCT\cite{Li_2017_NIPS}, Avatar-Net\cite{sheng2018avatar}.}
    \label{fig:additional_comparison1}
\end{figure}

\begin{figure}[t]\centering
    \includegraphics[width=\linewidth]{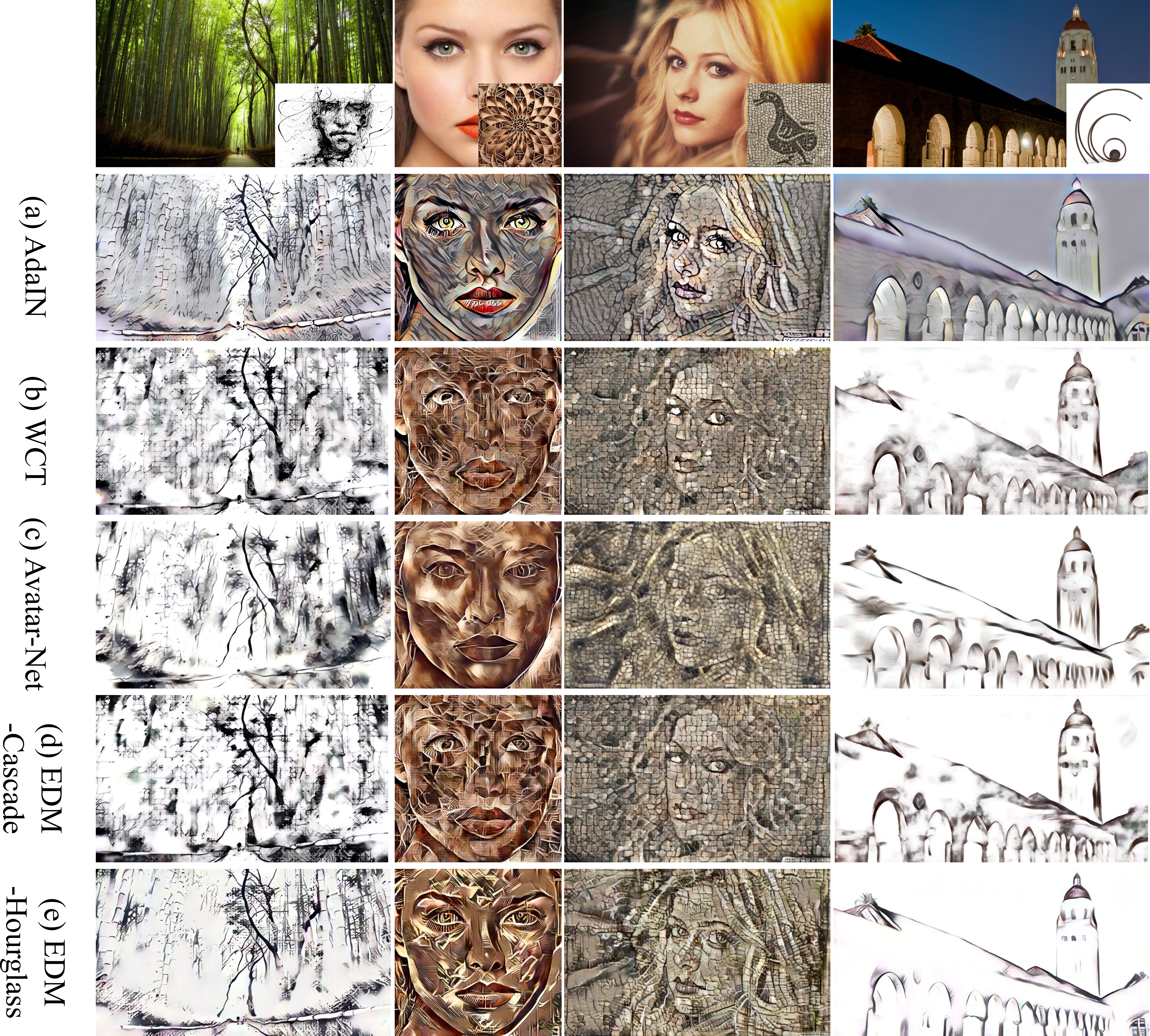}
    \caption{Additional comparison with other state-of-the-art: AdaIN\cite{Huang_2017_ICCV}, WCT\cite{Li_2017_NIPS}, Avatar-Net\cite{sheng2018avatar}.}
    \label{fig:additional_comparison2}
\end{figure}

We presented additional result images in fig.\ref{fig:additional_comparison1} and compared \ref{fig:additional_comparison2} with other methods.
As shown in the 1st column of fig.\ref{fig:additional_comparison1}, AdaIN, WCT, and Avatar-Net produce the artifact in the background, but EDM-Cascade generated images without any artifacts.
In the 2nd column, the EDM methods (EDM-Cascade, EDM-Hourglass) and Avatar-Net generate the images of varying stroke thickness that appears in the style image.
In the 3rd column, AdaIN produces the image that lacks style pattern. In contrast, WCT, Avatar-Net, and EDM-Cascade produce the overly stylized image, resulting in difficulty to recognize content. On the other hand, EDM-Hourglass generates a stylized image that has a proper balance between content and style.
In the 4th column, EDM-Cascade shows a fine image with better contrast than WCT.

In the 1st and 4th columns of fig.\ref{fig:additional_comparison2}, the EDM methods (especially EDM-Hourglass) shows the result images in which black and white are clearly distinguished like the style image.
In the 2nd column, EDM-Hourglass shows better contrast than WCT on the nose and eyes, stylizes well, and keeps the content a little better.
Lastly, in the 3rd column, while other methods failed to maintain the hair of the content, EDM-Hourglass succeeded making a good stylized image.

\clearpage
\section{Close-up comparison}
\begin{figure}[h]\centering
    \includegraphics[width=\linewidth]{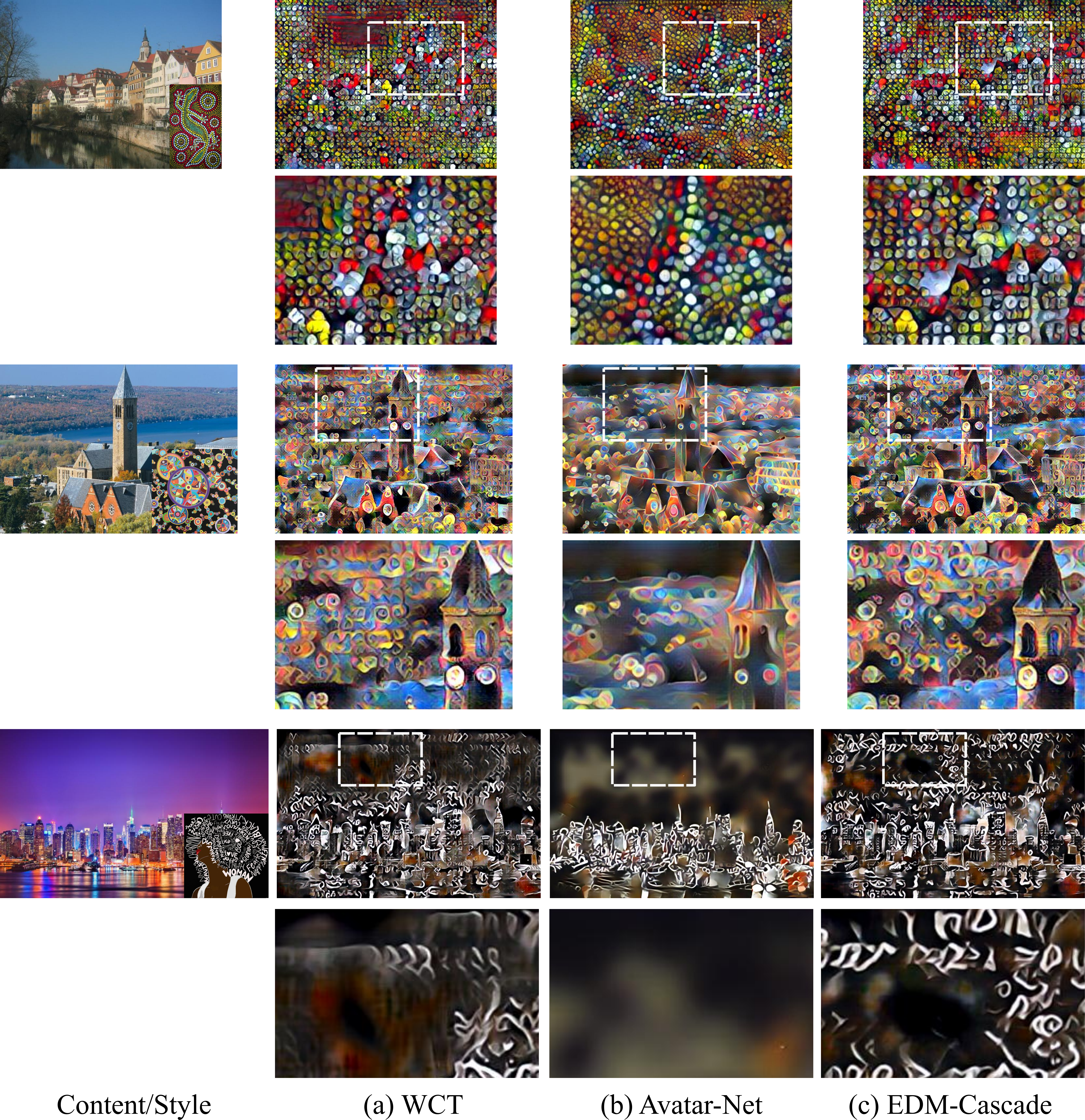}
    \caption{Close-up result. EDM-Cascade shows better results in the close-up part.}
    \label{fig:zoom}
\end{figure}

To show detail how the proposed EDM differs from WCT and Avatar-Net, the zoomed images are shown in fig.\ref{fig:zoom}.
The first result shows that EDM-Cascade, unlike WCT, transfers style onto the background and the foreground differently without artifacts on the background.
In the second, unlike WCT, EDM-Cascade evenly distributed the style pattern on the background, preventing the style pattern from being focused on one side.
In the third one, EDM-Cascade generated the continuous white line patterns of the style image on the result image without interruption.




\clearpage
%
%
\bibliographystyle{splncs04}
\bibliography{egbib}